\documentclass[runningheads]{llncs}
\usepackage[T1]{fontenc}
\usepackage[utf8]{inputenc}
\usepackage{placeins}
\usepackage{url}
\usepackage{graphicx}
\usepackage{amssymb}
\usepackage{booktabs}
\usepackage{xcolor}
\usepackage{multirow}
\usepackage{soul}
\usepackage{wrapfig}
\usepackage{subcaption}
\usepackage{adjustbox}
\usepackage{enumerate}
\usepackage{multicol}
\usepackage{array}
\usepackage{makecell}
\usepackage{enumitem}
\usepackage{verbatim}
\usepackage{changepage}
\usepackage{comment}
\usepackage{amsmath}
\usepackage{color}
\usepackage{url}
\usepackage{caption}

\usepackage{fancyvrb}
\definecolor{verbgray}{rgb}{0.3, 0.3, 0.3}
\DefineVerbatimEnvironment{myverbatim}{Verbatim}{fontfamily=cmtt,formatcom=\color{verbgray}}

\colorlet{dred}{red!50!black}
\colorlet{dblue}{blue!70!black}

\newcommand{\postcaption}{\vspace{0.75em}}

\newcommand{\tablefontsize}{\fontsize{8pt}{9pt}\selectfont}

\newcommand{\organismsubset}{\emph{organism}}
\newcommand{\nonorganismsubset}{\emph{non-organism}}
\newcommand{\notzsm}{--}

\newcommand{\ftyes}{$\star$}
\newcommand{\ftno}{}
\newcommand{\ftbasemodel}{}

\newcommand{\tblexplainmacs}{To measure compute cost, we calculate training MACs (multiply–accumulate operations).}
\newcommand{\tblgroups}{The table is grouped into training from scratch, finetuning, and analyzing components, the groups are separated by double horizontal lines. Each part is again split by single horizontal lines into groups of same model size or same component analysis.}
\newcommand{\tblexplain}{We mark the \textbf{best} and \underline{second best} result.}
\newcommand{\tblexplainnotzs}{We only compare zero-shot results and mark results as ``\notzsm{}'' if the model does not fulfill the zero-shot requirements.}
\newcommand{\tblexplainstar}{Models marked with $\star$ are finetuned.}

\newcommand{\entitynet}{EntityNet}
\newcommand{\entityneta}{}
\newcommand{\entitynetours}{EntityNet (ours)}
\newcommand{\datasetdatacomponeb}{DataComp-1B}
\newcommand{\datasetcctwelvemshort}{CC12M}

\newcommand{\minititle}[1]{\textbf{#1.}}

\newcommand{\writesubtitle}{
	\author{
    Simon Ging\thanks{Corresponding author. Email: \email{gings@cs.uni-freiburg.de}}
    \and Sebastian Walter%
    \and Jelena Bratulić%
    \and Johannes Dienert%
    \and Hannah Bast%
    \and Thomas Brox%
}
	\authorrunning{S. Ging et al.}
	\institute{University of Freiburg, Germany
    \email{\{gings,swalter,bratulic,dienertj,bast,brox\}@cs.uni-freiburg.de}
    \\Code and data:
    \texttt{\href{https://entity-net.github.io}{https://entity-net.github.io}}
    }
}

\newcommand{\rankone}[1]{\textbf{#1}}
\newcommand{\ranktwo}[1]{\underline{#1}}

\newcommand{\imagenetr}{\mbox{ImageNet-R}}
\newcommand{\imageneta}{\mbox{ImageNet-A}}
\newcommand{\imagenetvtwo}{\mbox{ImageNet-V2}}
\newcommand{\objectnet}{\mbox{ObjectNet}}

\usepackage{cite}
\usepackage{hyperref}

\usepackage[capitalize]{cleveref}
\crefname{section}{Sec.}{Secs.}
\Crefname{section}{Section}{Sections}
\Crefname{table}{Table}{Tables}
\crefname{table}{Tab.}{Tabs.}
\crefname{figure}{Fig.}{Figs.}
\Crefname{figure}{Figure}{Figures}

\newif\ifreview
\reviewfalse
\ifreview
	\usepackage{lineno}

	\linenumbers
\fi
\begin{document}
\def\SubNumber{083}
\def\GCPRTrack{Fast Review Track}
\title{
Using Knowledge Graphs to harvest datasets for efficient CLIP model training
}
\ifreview
\else
    \titlerunning{Using Knowledge Graphs to harvest datasets for efficient CLIP training}
    \writesubtitle{}
\fi
\maketitle

\begin{abstract}
Training high-quality CLIP models typically requires enormous datasets, which limits the development of domain-specific models -- especially in areas that even the largest CLIP models do not cover well -- and drives up training costs. This poses challenges for scientific research that needs fine-grained control over the training procedure of CLIP models. In this work, we show that by employing smart web search strategies enhanced with knowledge graphs, a robust CLIP model can be trained from scratch with considerably less data. Specifically, we demonstrate that an expert foundation model for living organisms can be built using just 10M images. Moreover, we introduce \mbox{EntityNet}, a dataset comprising 33M images paired with 46M text descriptions, which enables the training of a generic CLIP model in significantly reduced time.
\end{abstract}

\section{Introduction}\label{sec:intro}

\begin{figure}[t]
    \centering
    \includegraphics[width=1\linewidth]{fig-results1-wide.png}
    \caption{
    We demonstrate how to harvest datasets for training CLIP models with an improved quality-cost trade-off, for a generic (left) or an expert domain (right).
    }
    \label{fig:results1}
\end{figure}

Contrastive Language-Image Pretraining (CLIP) \cite{clip} has become a cornerstone for training Vision-Language Models (VLMs).
CLIP models learn high-quality visual embeddings and establish a link to the semantic level of brief text descriptions by training on pairs of images and their corresponding text descriptions collected from the web. The features and the link between images and text have been used directly for, e.g., zero-shot classification or text-to-image retrieval, and enable dialogues with visual input, such as in the LLaVA family of models~\cite{liu2023llava,liu2023improvedllava,liu2024llavanext}. The link can also be exploited in the opposite direction to enable text-conditional image generation, e.g., Stable Diffusion~\cite{sdxl}.

Training state-of-the-art CLIP models is computationally expensive. The original CLIP model has seen 12.8B image-text pairs, and later works have scaled this further~\cite{gadre2024datacomp,fang2024dfn}. 
This need for scale has limited most of the research to finetuning, which comes with reduced architectural flexibility and control over the data selection. It is particularly problematic for analytic research that demands full control over training to find causes of emergent behavior.

The effort to collect vast datasets is also a key bottleneck for building foundation models for expert domains. Although CLIP models are supposed to be generic and cover most of the world, they are not good enough for use in specific expert domains such as medicine or biology. Building foundation models for expert domains requires an efficient data collection process, taking into account the availability of fewer data samples in these domains.

Our goal is to tackle these challenges from the dataset side while keeping the CLIP algorithm fixed. This strategy is backed by recent literature. For example, Li~et~al.~\cite{li2024scaling} explored CLIP ``along three dimensions: data, architecture, and training strategies'' and they stress the ``significance of high-quality training data''.
For Large Language Models (LLMs), data curation was shown to reduce training time and model size, achieved through heavily filtered publicly available web data and synthetic data~\cite{phi3}.
With the dataset creation process, we aim (1) for improved performance in the expert domain of living organisms, in order to demonstrate the creation of expert foundation models; and (2) we aim for  a good trade-off between training efficiency and model performance on the broad domain of the visual world, in order to enable compute-efficient from-scratch analysis of fully functional CLIP models. 

We built a dataset we call \emph{EntityNet}, where we leveraged knowledge graphs and targeted web image search. Specifically, from the knowledge graphs Wikidata and WordNet, we collected 135k entities (e.g. \emph{eagle}) as well as their aliases and descriptions.
We extracted entity attributes from Wikidata related to color, partonomy, behavior, and other aspects, and used them to guide an LLM in generating entity-attribute queries for image search. For example, from the entity \emph{plastic} and the attribute \emph{small} we generated the query \emph{small plastic item}.

The resulting EntityNet consists of 33M images paired with 45M alt texts and 613k text labels from the \mbox{knowledge} graphs. 
The dataset is partitioned into a subset of 10M images of living organisms, capturing high-quality visual and semantic information about the taxonomy of animals, plants, and funghi, as well as a subset of 23M images covering a wide range of categories, such as tools, geographical features, materials, and buildings.
Notably, from this process we obtain not only alt texts, but also a link back to the knowledge graph information that was used to create the search query for a given image. We show that this information can be used during training to achieve better performance than by training on alt texts alone.
The method of creating our dataset is largely generic and can be applied to other knowledge graphs.

Training on this dataset, we obtain a foundation model that is both specialized on the target expert domain and is also able to understand the overall visual world.
In our domain-specific evaluations on iNaturalist and RareSpecies, the model demonstrates robust generalization and clearly surpasses CLIP models trained on much more data (\Cref{fig:results1}).
On ImageNet, we demonstrate our dataset to be highly compute efficient and to achieve a performance comparable to models trained 20x longer (\Cref{fig:results1}).

\begin{itemize}[leftmargin=1em]
\item 
We propose a method to automatically create a vision-language dataset based on a given knowledge graph and an image search engine.
\item We apply this method to create the \emph{EntityNet} dataset, consisting of 33M images paired with 45M alt texts and supplementary text information from the knowledge graphs.
\item We train an expert CLIP model for living organisms on a single 8xL40S machine from scratch in 55 hours. This \emph{EntityNet-CLIP} is highly specialized in the target expert domain of living organisms, \emph{and} comparably strong on ImageNet.
\item We evaluate our model and a suite of other CLIP models for object classification, image retrieval, and domain shift robustness.
In the expert domain of animals and plants, our model achieves higher performance than models with orders of magnitude more parameters or training data. It is also much stronger than CLIP models that specialize only for this domain.
In the generic domain, our model performs remarkably well given the low amount of compute required to train it.
\end{itemize}
\begin{figure}[t]
    \centering
    \includegraphics[width=0.9\linewidth]{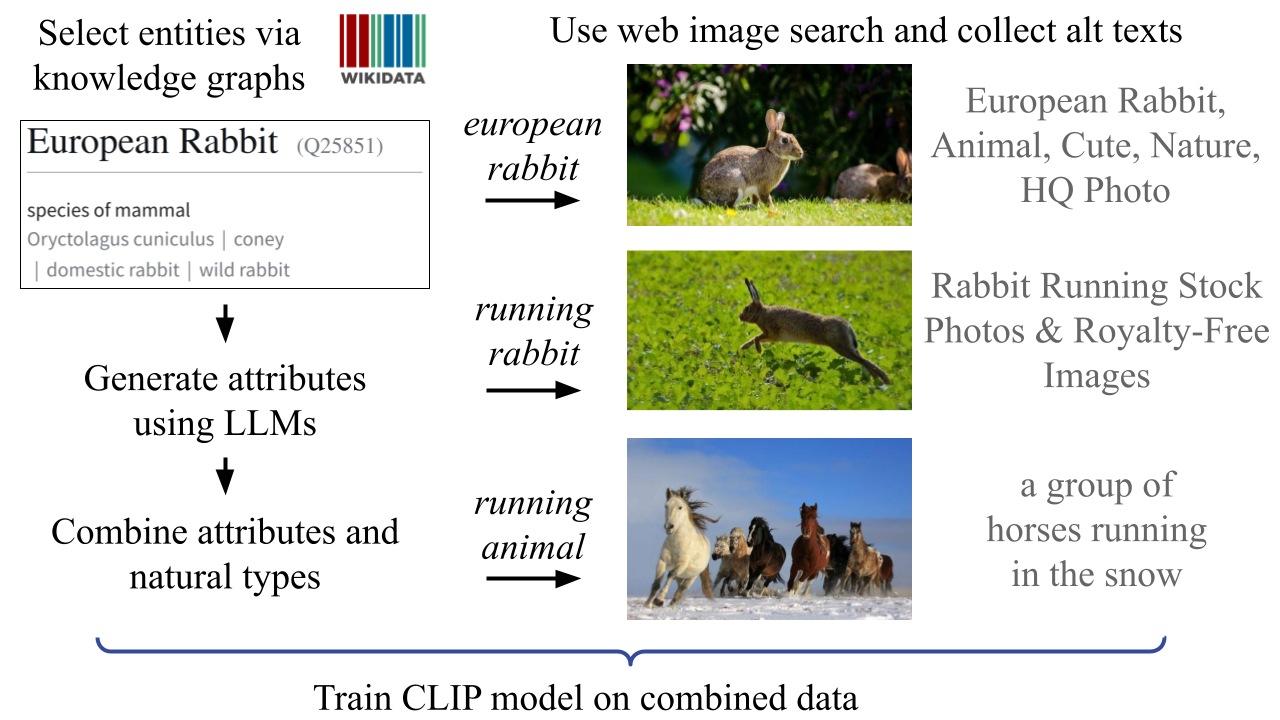}
    \caption{
We create a dataset for vision-language pretraining: First, we extract entities from knowledge graphs, then generate attributes and natural types for them. We search for different combinations of entities, attributes, and types in image search engines, and collect alt texts for each image. Finally, we train our model on the combined data.    
    }
    \label{fig:datasetprocess}
\end{figure}

\section{Related work}\label{sec:related}

\minititle{Datasets}
Many recent studies have investigated methods for building large-scale datasets for multimodal training.
\mbox{Radford}~et~al.~\cite{clip} trained the original CLIP model on a private dataset of 400M images using image-text pairs with text derived from Wikipedia and WordNet terms. Schuhmann~et~al.~\cite{schuhmann2021laion} further built the publicly available LAION-400M dataset
by filtering HTML data from Common Crawl ~\cite{rana2010commoncrawl} based on the similarity estimated by the CLIP model. In a follow-up work, they~\cite{schuhmann2022laion} scaled their approach one order of magnitude with the multilingual LAION-5B dataset.
Xu~et~al.~\cite{xu2024demystifying} sought to replicate the original CLIP's data curation approach. %
Gadre~et~al.~\cite{gadre2024datacomp} proposed DataComp, a filtering challenge containing up to 13B image-text pairs from CommonCrawl, and a baseline DataComp-1B dataset. It contains 1.4B pairs filtered with a combination of CLIP score and clustering to match ImageNet~\cite{imagenet} training examples. 
Fang~et~al.~\cite{fang2024dfn} trained a Data Filtering Network on 357M human-verified image-text pairs, which they used to filter 42B candidates into the DFN-5B dataset and use that dataset to train the current top model of the OpenCLIP leaderboard~\cite{ilharco2021openclip}.
These large datasets have largely supplanted smaller ones like ConceptualCaptions12M (CC12M)~\cite{changpinyo2021cc12m}, relying on unimodal heuristics, and Yahoo Flickr Creative Commons 15M (YFCC15M), a derived subset of 15M image-text pairs from Flickr~\cite{bart2016yfcc100m}. 
While many prior works have focused on scaling up multimodal datasets and models, we aim to improve research on high-quality CLIP models when data and compute efficiency is essential, such as setting up a CLIP model for an expert domain or for scientific analysis of CLIP training. 

Stevens~et~al.~\cite{bioclip} curated the TreeOfLife-10M dataset from biological sources~\cite{EOL,inaturalist21,bioscan1m} %
to train BioCLIP, a model for organismal biology. They evaluated it on RareSpecies, a benchmark of 400 species not seen during training. While BioCLIP leverages domain-specific biological knowledge, we propose a dataset construction method that generalizes to arbitrary domains using knowledge graphs.

\minititle{Training algorithms}
Several works have investigated algorithmic improvements to CLIP. 
Li~et~al.~\cite{li2023reclip} simply train and fine-tune with different image resolutions, while Li~et~al.~\cite{li2023clipav2} suggests masking parts of the image to reduce computation. Zhai~et~al.~\cite{zhai2023siglip} propose a sigmoid loss which reduces the computational load especially in big distributed settings. They follow up~\cite{tschannen2025siglip} by extending the training objective using multiple previously developed techniques, including captioning-based pretraining~\cite{wan2024locca}, self-distillation~\cite{naeem2024silc} and online data curation~\cite{udandarao2024active} 
into a unified training strategy. %
Vasu~et~al.~\cite{mobileclip2024} improve learning efficiency with synthetic captions created by an image captioning model and an ensemble of CLIP teachers to train their model.
Chen~et~al.~\cite{chen2024vitamin} evaluate vision encoder choices and design a hybrid architecture that improves over vanilla vision transformers (ViT) based CLIP models.
These algorithmic improvements are orthogonal to our contribution. In this work, we fix the algorithm and architecture to enable a fair comparison with ViT-based CLIP baselines.

Li~et~al.~\cite{li2024scaling} analyze scaling effects across data, architecture, and training strategies, showing that huge models require larger datasets, and data quality plays a crucial role. They create improved datasets by filtering the \mbox{3.4B~WebLI} dataset~\cite{chen2023pali} with CLIP, while we pursue a different dataset collection process.

\section{Dataset creation}\label{sec:dataset-creation}

Our dataset creation process consists of four steps: entity extraction, attribute generation, query building, and image search. This process is generally applicable to all visual domains covered by the underlying knowledge graph.
We construct a dataset covering most visual entities in our world, additionally focusing on animals and plants, referred to as the 
\organismsubset{} subset.
See \Cref{fig:datasetprocess} for the dataset creation process and \Cref{tab:entitynet-ent-att} for examples of entities and attributes.

\subsection{Entity extraction}\label{subsec:entities}

A high-quality list of visual entities forms the basis for our dataset, built from the Wikidata knowledge graph~\cite{wikidata} and utilizing the hierarchical structure provided by the \href{https://www.wikidata.org/wiki/Property:P279}{\emph{subclass of}} relation within Wikidata. 
For example, the entity \href{https://www.wikidata.org/wiki/Q144}{\emph{dog}} is a subclass of the \href{https://www.wikidata.org/wiki/Q39201}{\emph{pet}} entity, which in turn is a subclass of \href{https://www.wikidata.org/wiki/Q622852}{\emph{domesticated animal}}. 
This hierarchy enables easy collection of entities related to a super-entity.
First, we manually build a list of 21 super-entities that cover most physical and visual entities in Wikidata. For the \organismsubset{} subset, the super-entities are just \href{https://www.wikidata.org/wiki/Q729}{\emph{animal}} and \href{https://www.wikidata.org/wiki/Q756}{\emph{plant}}. Examples of \nonorganismsubset{} super-entities include \href{https://www.wikidata.org/wiki/Q2095}{\emph{food}}, \href{https://www.wikidata.org/wiki/Q41176}{\emph{building}}, or \href{https://www.wikidata.org/wiki/Q39546}{\emph{physical tool}}, with all super-entities listed in the supplementary material.
Next, we extract all entities from Wikidata linked to at least one of the super-entities through the \emph{subclass of} relation. 
For animals and plants, Wikidata also models their biological taxon hierarchy via the \href{https://www.wikidata.org/wiki/Property:P171}{\emph{parent taxon}} relation. Because the taxon hierarchy substantially increases the coverage of our \organismsubset{} subset, we use it together with the regular \emph{subclass} hierarchy to extract entities.
We exclude named entities (e.g., specific persons), as Wikidata models these via the \emph{instance of} relation; we focus solely on the \emph{subclass of} and \emph{parent taxon} relations.
For every entity, we also download its name, description, aliases, and its number of Wikimedia sitelinks \footnote{The number of Wikimedia sitelinks is a commonly used and high-quality proxy for the popularity of an entity \cite{peshterliev2020self}.} as additional information.
Finally, we apply two filtering steps: First, we remove all entities with a sitelink count below a predefined threshold, eliminating very rare or low-quality entities unlikely to produce strong search results. 
We then use a LLM to filter out any remaining non-visual entities.

For our expert domain, the \organismsubset{} subset, we also add all nouns from WordNet~\cite{fellbaum1998wordnet} that are a subclass of the \emph{living thing} node, excluding humans, named entities and entities that cannot be seen with the bare eye, e.g., microorganisms.
Finally, we employ heuristic methods to detect and remove potentially offensive entities via a profanity filter.

\begin{table}[t]
\caption{
\textbf{Top:} Examples of entities
and additional information as extracted from the Wikidata knowledge graph.
\textbf{Bottom:}
Examples of attributes and corresponding search queries for different entities as generated by the LLM.
\postcaption{}
}\label{tab:entitynet-ent-att}
{
\tablefontsize
\setlength{\tabcolsep}{4pt}
\begin{tabular}{llrl}
\toprule
Entity & Description & Sitelinks & Aliases \\
\midrule
\href{http://www.wikidata.org/entity/Q19939}{tiger} &
species of big cat & 216 & tigress, tigers, Panthera tigris \\
\href{http://www.wikidata.org/entity/Q366134}{chest	} &
box-shaped type of furniture & 51 & coffer, kist \\
\href{http://www.wikidata.org/entity/Q1072763}{muscle car} &
type of high-performance car & 30 & high performance car \\
\toprule
Entity & Category &  Attribute & Search query \\
\midrule
\href{https://www.wikidata.org/entity/Q8063}{rock} & Pattern and texture & porous & porous rock \\
\href{https://www.wikidata.org/entity/Q18498}{wolf} & Environment & snow & wolf in the snow \\
\href{http://www.wikidata.org/entity/Q699405}{residence} & Parts & arches & arches in residence architecture \\
\href{https://www.wikidata.org/entity/Q23400}{garlic} & Shape and size & big & big garlic bulb \\
\href{https://www.wikidata.org/entity/Q131596}{farm} & Other & tourist & tourists visiting a farm \\
\href{http://www.wikidata.org/entity/Q190868}{boot} & Color & multicolor & multicolored boots \\
\bottomrule
\end{tabular}
}
\end{table}

\subsection{Attribute generation}\label{subsec:attributes}

Besides searching for the entities directly, we also aim to search for variations of them in different contexts, by combining them with attributes.
We manually define 6 visual attribute categories we want to search for: \emph{Color}, \emph{Pattern and texture}, \emph{Parts}, \emph{Shape and size}, \emph{Environment}, and \emph{Other}.
We extract potential attributes for each entity from the Wikidata knowledge graph and prompt an LLM\footnote{We use three LLMs and merge their generated attributes: Qwen2.5 7B~\cite{qwen2.5}, OpenAI GPT-4o, and OpenAI GPT-4o mini (both accessed via API at \href{https://platform.openai.com/}{platform.openai.com})} with this entity and attribute information to generate a list of visual attributes. We first considered generating attributes without categories, however, the results lacked diversity, and adding categories improves the variety of attributes.
For each attribute we also generate a search query combining the attribute itself with the corresponding entity.
We generate between 1 and 10 attributes per category and generate them for the most popular entities only, as image search engines fail to respect attributes in search queries for rare entities, where they often even struggle to return good results for the entity alone.

\subsection{Query building}

For the entities themselves, we use their names and aliases as search queries. We search entity-attribute combinations using the search queries generated by the LLM. We then create additional queries based on the attributes: First, we determine the entity's natural type -- the super-entity a human would most likely associate with it, e.g., \emph{bird} for \emph{eagle}, or \emph{clothing} for \emph{hat}. It is neither too general nor too specific, and can typically help disambiguate entities sharing the same name. %
We use an LLM to select the most fitting super-entity from an entity's super-class hierarchy as its natural type and generate a brief description explaining why this type is appropriate. %
The description is used during training as a potential text label. We then replace entity mentions in the attribute search queries with their natural types. For example, the attribute query \emph{eagle in its nest} may turn into \emph{bird in its nest}, or \emph{black BMW M4} into \emph{black car}.

\subsection{Image search and filtering}\label{subsec:imagesearch}

We execute our search queries using the image search APIs of Bing and Google. Initial tests on the \organismsubset{} subset revealed  Bing's search results to be of much higher quality at a lower cost, so we rely solely on the Bing API for all other queries. The image search APIs also provide the URL for the website hosting the image, which we use to collect alt texts from the HTML image tag.
After downloading images and alt texts, we perform the following postprocessing steps.

Similar to Changpinyo~et~al.~\cite{changpinyo2021cc12m}, we apply \emph{relaxed filtering heuristics}.
We do not use multimodal filtering, but rely on search engines to provide image-text correspondences. We remove JSON-like and too long text.
We also remove images with an aspect ratio of more than 4 or covering less than 4096 pixels.

We deduplicate all downloaded images using the \mbox{Self-Supervised} Descriptor for Image Copy Detection method (SSCD)~\cite{pizzi2022sscd}. For duplicates, we retain the largest image and collect all unique alt texts and related entities from the duplicates.
Deduplication increases the dataset diversity per sample, since the domain coverage stays the same, while the number of samples decreases.
We also remove images that appear in any evaluation dataset using the same SSCD method.

Our final dataset comprises approximately 33M images and 45M alt texts, obtained from 416k queries. This amounts to 79 images per query and 1.4 alt texts per image on average. The total cost for all image search API calls was around 10,000\$.
See \Cref{tab:entitynet-stats} for an overview over our dataset.

\begin{table}[t]
\caption{Details of our EntityNet dataset. We show the number of unique elements for each column, e.g. the number of images after deduplication or all unique entity aliases in the respective sets.\postcaption{}
}\label{tab:entitynet-stats}
{\tablefontsize
\setlength{\tabcolsep}{2pt}
\begin{tabular}{l|rrrrrrl}
\toprule
Query set & Images & Queries & Entities & Aliases & Attributes & Alt texts & Example query \\
\midrule
World entity & 23M & 158k & 74k & 101k & - & 23M & ship \\
+ attribute & 19M & 139k & 6k & 16k & 20k & 16M & small handbag\\
Living entity & 9M & 72k & 63k & 51k & - & 8M & kohlrabi \\
+ attribute & 9M & 53k & 1k & 3k & 5k & 6M & tropical plant \\
\midrule
All & 33M & 416k & 135k & 149k & 23k & 45M & -\\
\bottomrule
\end{tabular}}
\end{table}

\section{Experimental setup}\label{sec:experimental-setup}

\subsection{Training} 

We trained all models with the standard CLIP loss~\cite{clip}, using a batch size of 8,192 for pretraining and 32,768 during finetuning, along with random resized crop augmentation. We sampled text labels from both the image alt texts and the knowledge graph. For each image, 50\% of the time, we chose a random alt text, and 50\% of the time, we chose randomly between search queries, aliases, or descriptions of the corresponding entity. We trained all models for 18 epochs.
Training on 33M images takes $\sim$55 hours on 8 L40s GPUs (48GB VRAM per GPU).
Our training code is based on OpenCLIP~\cite{ilharco2021openclip}. Further training details and text sampling examples are in the supplementary material.

\subsection{Evaluated models}

On our EntityNet dataset, we trained ViT CLIP models of size B-32 and B-16 from random initialization. For a comparison with a similarly sized dataset, we also trained models on CC12M by downloading all available URLs, and then detecting and removing duplicates relative to the evaluation datasets using the same procedure as detailed in \Cref{subsec:imagesearch}, obtaining 9.3M images.
We finetuned B-32 and B-16 CLIP models trained on \datasetdatacomponeb{} on our dataset to compare finetuning and pretraining performance.
We also evaluate the original \emph{OpenAI CLIP}~\cite{clip}, models pretrained on \emph{DataComp-M/L/1B}, \emph{CommonPool-M/L}~\cite{gadre2024datacomp}, and DFN-5B~\cite{fang2024dfn},
as well as the biological domain expert model BioCLIP~\cite{bioclip}, a ViT-B-16 CLIP model finetuned from OpenAI-CLIP on the TreeOfLife-10M dataset.

\subsection{Object classification evaluation}\label{sec:objclseval}

To test the VLMs on object classification, we use the same procedure as CLIP~\cite{clip}, see the supplementary material for a detailed description.
We evaluate all models on just encoding the class name, and on using the average embedding of the 80 context prompts that the CLIP authors used for ImageNet, and report the higher top-1 accuracy.
For zero-shot object classification, we require models not to have been trained on the training set of the benchmark, to test ``generalization to unseen datasets''~\cite{clip}

\minititle{Benchmarks in the generic domain}\label{sec:method_datasets} We evaluate on ImageNet~\cite{imagenet}, a popular image classification benchmark~\cite{ilsvrc15}. We use the ILSVRC2012 validation set, which contains 50,000 images from 1,000 classes. The classes include simple objects, such as \emph{park bench}, as well as more fine-grained labels like 23 types of terrier dogs, e.g., \emph{Staffordshire Bull Terrier}.
We further evaluate the robustness under distribution shifts on \imageneta{}~\cite{hendrycks2021natural}, \imagenetr{}~\cite{hendrycks2021many}, ImageNet-Sketch~\cite{wang2019learning}, \imagenetvtwo{}~\cite{recht2019imagenet}, and \objectnet{}~\cite{barbu2019objectnet} as proposed by Taori~et~al.~\cite{taori2020measuring}.
\imageneta{} contains 7500 samples of 200 ImageNet classes. The samples were adversarially filtered to make \mbox{ResNet-50s} misclassify them, providing a more challenging test. \imagenetr{} contains 30000 renditions, such as paintings or embroidery, of 200 ImageNet object classes. ImageNet-Sketch contains 50000 sketches, covering 200 ImageNet classes. \imagenetvtwo{} replicates the original ImageNet generation process, providing an additional 10000 test images. The object-centered \objectnet{} contains 18574 images from 113 ImageNet classes with control over background, rotation and viewpoint.

\minititle{Benchmarks in the expert domain}
We evaluate our models on iNaturalist~2021~\cite{inaturalist21}, a fine-grained species classification benchmark that contains 100k images in the validation set of 10k different species. Similar to Parashar~et~al.~\cite{parashar2023prompting}, we report the best results after testing on both the English common name and the Latin taxon name. 
We further test on the Caltech-UCSD Birds (CUB)~\cite{cub} dataset, which contains 5,794 images of birds in the original author's test set, each annotated as one of 200 fine-grained bird species, e.g., \emph{grasshopper sparrow}.
Additionally, we evaluate on the Rare Species benchmark proposed by Stevens~et~al.~\cite{bioclip}, comprising 400 species with 30 images each, specifically tailored to assess generalization to unseen taxa.
To comply with the benchmark requirements of not seeing the testing 400 species during training, we exclude all entities and queries from our dataset that appear in RareSpecies, using substring matching.
As class names, we evaluate all text types proposed by Stevens~et~al.~\cite{bioclip}: combinations of the Latin taxonomy and the English common name. Same as in \Cref{sec:objclseval} we evaluate on both the CLIP ImageNet prompt and no prompt, and report the better of both accuracies.

\subsection{Retrieval evaluation}\label{sec:attrclseval}

We evaluate the COCO Karpathy test split~\cite{karpathy2015deep}, a subset of 5000 samples from the MS-COCO \cite{lin2014microsoft} dataset paired with 5 texts each. We also evaluate the 1000 samples in the Karpathy test split of Flickr30k~\cite{young2014image} annotated with 5 texts per image, as well as on the 3600 image-text pairs in XM3600~\cite{thapliyal2022crossmodal}.
We report the average of image-to-text and text-to-image recall@1 over all datasets.

\section{Results}\label{sec:results}

\begin{figure}[t]
    \centering
    \includegraphics[width=1\linewidth]{fig-results2-wide.png}
    \caption{
    Results on image retrieval and distribution shift robustness on ImageNet.
    }
    \label{fig:results2}
\end{figure}

\begin{table}[t]
\caption{
Results for training CLIP B-32 and B-16 on our EntityNet dataset from scratch.
\tblexplain{}
\tblexplainmacs{}
\tblexplainnotzs{}\postcaption{}
}\label{tab:results-scratch}
{\tablefontsize
\setlength{\tabcolsep}{2.8pt}
\begin{tabular}{llrr|rrr|rrrrrrrr}
\toprule
Arch. & Dataset & MACs  & Images in & Image- & Retrie- & Distr. & iNat. & CUB & Rare \\
&& (1e18)& dataset (M) & Net & val & shift & 2021 & & Species \\
\midrule
B-32 & \datasetcctwelvemshort{}  & 3.7 & 9.3 & 28.6 & 25.6 & 18.3 & 0.7 & 9.2 & \notzsm{} \\
B-32 & CommonPool-M  & 2.9 & 128.0 & 27.2 & 20.2 & 19.8 & 0.8 & 10.1 & \notzsm{} \\
B-32 & DataComp-M  & 2.9 & 14.0 & 29.7 & 19.5 & 20.5 & 1.0 & 16.8 & \notzsm{} \\
B-32 & OpenAI  & 288.6 & 400.0 & \ranktwo{63.4} & \ranktwo{49.6} & \ranktwo{48.7} & 7.4 & 51.8 & \notzsm{} \\
B-32 & DataComp-1B  & 295.4 & 1400.0 & \rankone{69.2} & \rankone{54.0} & \rankone{56.3} & \ranktwo{12.6} & \ranktwo{73.8} & \notzsm{} \\
\midrule
B-32 & \entitynetours{}  & 13.1 & 32.7 & 61.5 & 37.2 & 41.0 & \rankone{26.1} & \rankone{79.5} & \rankone{42.7} \\
\toprule
B-16 & BioCLIP & 61.3 & 10.4 & 18.6 & 0.8 & 15.4 & \notzsm{} & 78.1 & \ranktwo{38.1} \\
B-16 & CommonPool-L  & 78.2 & 1280.0 & 57.8 & 45.6 & 47.0 & 4.1 & 35.1 & \notzsm{} \\
B-16 & DataComp-L  & 78.2 & 140.0 & 63.1 & 49.4 & 51.1 & 6.1 & 48.1 & \notzsm{} \\
B-16 & DataComp-1B  & 791.4 & 1400.0 & \rankone{73.5} & \rankone{57.4} & \rankone{64.4} & \ranktwo{15.3} & \ranktwo{79.0} & \notzsm{} \\
B-16 & OpenAI  & 784.6 & 400.0 & \ranktwo{68.3} & \ranktwo{52.1} & \ranktwo{58.6} & 9.2 & 56.1 & \notzsm{} \\
\midrule
B-16 & \entitynetours{}  & 36.0 & 32.7 & 66.2 & 39.8 & 47.4 & \rankone{32.0} & \rankone{85.3} & \rankone{47.1} \\
\toprule
L-14 & OpenAI  & 3328.4 & 400.0 & 75.5 & 54.3 & 71.4 & 12.0 & 62.9 & \notzsm{} \\
L-14 & DataComp-1B  & 3338.6 & 1400.0 & 79.2 & 61.8 & \ranktwo{74.9} & 21.1 & 85.5 & \notzsm{} \\
L-14 & DFN-2B  & 3338.6 & 2000.0 & \ranktwo{81.4} & \ranktwo{64.2} & 74.8 & 21.6 & \ranktwo{86.5} & \notzsm{} \\
H-14 & DFN-5B  & 22164.0 & 5000.0 & \rankone{83.4} & \rankone{68.7} & \rankone{76.3} & \ranktwo{25.1} & \rankone{88.1} & \notzsm{} \\
\bottomrule
\end{tabular}
}
\end{table}

\begin{table}[t]
\caption{
Results for finetuning the DataComp-1B CLIP model on EntityNet.
\tblexplain{}
\tblexplainmacs{}\postcaption{}
}\label{tab:results-ft}
{\tablefontsize
\setlength{\tabcolsep}{4pt}
\begin{tabular}{llrr|rrr|rrrrrrrr}
\toprule
Arch. & Dataset & MACs  & Images in & Image & Retrie- & Distr. & iNat & CUB
\\
& & (1e18) & dataset (M) & -Net & val & shift & 2021 & 
\\
\midrule
B-32 & DataComp-1B \ftbasemodel{} & 295.4 & 1400.0 & \ranktwo{69.2} & \rankone{54.0} & \rankone{56.3} & 12.6 & 73.8 \\
B-32 & \entitynet{}  & 13.3 & 32.7 & \rankone{69.5} & \ranktwo{50.8} & \ranktwo{53.3} & \ranktwo{29.5} & \ranktwo{83.3} \\
B-32 & \entityneta{}Only organisms & 4.2 & 10.2 & 48.2 & 31.2 & 33.3 & \rankone{37.0} & \rankone{87.0} \\
\midrule
B-16 & DataComp-1B \ftbasemodel{} & 791.4 & 1400.0 & \rankone{73.5} & \rankone{57.4} & \rankone{64.4} & 15.3 & 79.0 \\
B-16 & \entitynet{}  & 36.1 & 32.7 & \rankone{73.5} & \ranktwo{52.2} & \ranktwo{61.0} & \ranktwo{34.9} & \ranktwo{86.5} \\
B-16 & \entityneta{}Only organisms  & 11.3 & 10.2 & 51.4 & 34.8 & 39.2 & \rankone{42.7} & \rankone{90.3} \\
\bottomrule
\end{tabular}
}
\end{table}

\begin{table}[t]
\caption{
Analysis of performance when varying dataset composition, text sampling and dataset size.
\tblexplain{}
\tblexplainmacs{}
\postcaption{}
}\label{tab:results-abl}
{
\setlength{\tabcolsep}{3.2pt}
\tablefontsize
\begin{tabular}{llrr|rrr|rrrrrrrr}
\toprule
Arch. & Dataset & MACs  & Images in & Image- & Retrie- & Distr. & iNat. & CUB & Rare \\
&& (1e18)& dataset (M) & Net & val & shift & 2021 & & Species \\
\midrule
B-32 & Everything & 13.1 & 32.7 & \rankone{61.5} & \rankone{37.2} & \rankone{41.0} & \ranktwo{26.1} & 79.5 & \rankone{42.7} \\
B-32 & No organisms & 9.0 & 22.5 & 39.2 & \ranktwo{32.1} & 28.0 & 0.8 & 6.2 & 6.9 \\
B-32 & Only organisms & 4.1 & 10.2 & 36.0 & 16.5 & 21.0 & \rankone{28.6} & \rankone{83.2} & \ranktwo{42.0} \\
B-32 & No attributes & 8.7 & 21.8 & \ranktwo{54.8} & 28.6 & \ranktwo{33.8} & 25.6 & \ranktwo{79.7} & 39.2 \\
\midrule
B-32 & 50\% alt text & 13.1 & 32.7 & \rankone{61.5} & \ranktwo{37.2} & \rankone{41.0} & \rankone{26.1} & \rankone{79.5} & \rankone{42.7} \\
B-32 & 100\% alt text & 13.1 & 32.7 & \ranktwo{59.1} & \rankone{38.3} & \ranktwo{38.1} & 22.9 & \ranktwo{78.8} & \ranktwo{39.7} \\
B-32 & 0\% alt text & 13.1 & 32.7 & 55.7 & 13.5 & 35.5 & \ranktwo{24.2} & 78.1 & 29.7 \\
\midrule
B-32 & Full size & 13.1 & 32.7 & \rankone{61.5} & \rankone{37.2} & \rankone{41.0} & \rankone{26.1} & \rankone{79.5} & \rankone{42.7} \\
B-32 & 1/2 size & 6.6 & 16.4 & \ranktwo{54.1} & \ranktwo{30.3} & \ranktwo{33.6} & \ranktwo{20.1} & \ranktwo{74.1} & \ranktwo{36.6} \\
B-32 & 1/4 size & 3.3 & 8.2 & 45.2 & 23.5 & 25.7 & 13.2 & 64.9 & 28.0 \\
B-32 & 1/8 size & 1.6 & 4.1 & 33.3 & 16.6 & 17.7 & 7.3 & 47.8 & 19.4 \\
B-32 & 1/16 size & 0.8 & 2.0 & 19.9 & 9.4 & 10.0 & 2.8 & 27.4 & 11.8 \\
\bottomrule
\end{tabular}
}
\end{table}

We evaluate CLIP \textbf{pretrained from scratch} on our EntityNet dataset and CLIP models trained on other datasets. In Figures \ref{fig:results1} and \ref{fig:results2} we contrast the effectiveness of models with their training cost. We show the results in detail in \Cref{tab:results-scratch}.
In the generic domain, our models surpass others trained on similarly sized datasets while achieving comparable performance on object classification with models trained 20x longer.
On image-text retrieval, our model performs similarly to models trained on the same amount of compute. While our pipeline creates a dataset efficient for understanding objects and their properties, understanding complex scenes still requires learning mainly from the alt texts more than from objects and attributes.
In the expert domain, we outperform even the largest CLIP models on the challenging iNaturalist 2021 dataset, which requires classifying images among 10k fine-grained species.
Our model also excels on CUB by distinguishing 200 bird species better than all other CLIP models of the same size.
Further, when compared to the expert model \emph{BioCLIP}, explicitly trained for organismal biology at a similar training cost, our model demonstrates superior performance.
On the Rare Species benchmark, our model outperforms \emph{BioCLIP} on unseen species, showing the effectiveness of our dataset collection method over a manually designed living organism training set.

We investigate improving existing CLIP models via \textbf{finetuning} in \Cref{tab:results-ft}.
The results show that our dataset can be leveraged to create expert CLIP models that outperform both the base model and our model pretrained from scratch on the expert domain. This improvement comes at the cost of trading off some capabilities in the other domains. When finetuning only on the expert domain, we trade off more capabilities, yet obtain even stronger experts.

We further validate and verify our design choices through a \textbf{component analysis} in \Cref{tab:results-abl}.
Training separately on the generic and the domain expert part of our dataset reveals that, while the best generic model emerges from training on everything, a slightly better expert model is the result of training only on the expert domain (first table segment). However, generalization to unseen species slightly benefits when training on the full dataset, showing that our generic domain data can enhance generalization capabilities within the expert domain. We also observe that generating and downloading attribute queries contributes to improved performance of the pretrained model.

In the second segment of the table, we evaluate the mixture of alt text and knowledge graph labels used during training. Notably, both training only on alt texts or only on knowledge graph labels mostly performs worse than our \mbox{50-50} mix. The exception is image-text retrieval, where training fully on alt text performs slightly better. Potentially, the knowledge graph labels are less useful for learning the matching between longer text queries and images, and more useful for learning fine-grained object classification.

Finally, we reduce the scale of our dataset by powers of two. While the model performance expectedly drops with reduced dataset size, the efficiency of our dataset per datapoint stays high, with the model still reaching 33\% accuracy on ImageNet with only 4M images.

\section{Limitations}\label{sec:limitation}

The proposed data harvesting approach assumes that there is a knowledge graph for the target domain and that there is a searchable database with noisy pairing of images and text.
However, 
knowledge graphs exist in many domains, e.g.,
\emph{UniProt}~\cite{uniprot} with 246M protein sequence records or
\emph{AgriKG}~\cite{agrikg} with 150k agricultural entities.
Also, if no image search engine is available for the given domain, but a large amount of image-text data exists, pairs can be found by searching for the queries via substring matching in the image-text pairs.

Another limitation is the small, but significant drop in performance on image-text retrieval and classifying ImageNet distribution shifts in the generic domain, when finetuning a large model with \entitynet{}. 
First, our dataset by design has a strong focus on the expert domain and trades off some performance in the generic domain during finetuning. 
Second, our search pipeline finds many clean object-centric images and annotates them with entity information, which tremendously helps understanding object semantics, but to improve efficiency on image-text retrieval in a similar way, one needs to tackle the quality of alt texts and their alignment to the images~\cite{Xu2024AltogetherIC}. Finally, we focused on searching photos, which explains slightly lower accuracy when classifying paintings and sketches -- the \entitynet{} dataset simply contains a lower percentage of such types of images than datasets like \mbox{CommonPool}.

\section{Conclusions}\label{sec:conclusions}

We demonstrated how to use knowledge graphs to harvest datasets that are efficient for training CLIP models. Our strategy allows to create an expert domain dataset with little manual effort, enabling the development of CLIP models that significantly outperform standard models in the expert domain.
The expert domain dataset can be used for training a model from scratch or for finetuning an existing vanilla model. 
The substantial size and diversity of the expert domain dataset ensures that the good generalization properties of CLIP exist also in the expert domain, in contrast to training with an over-specialized expert dataset.   

Furthermore, we demonstrated that the proposed harvesting strategy is also viable to create a common domain dataset, which allows us to achieve a better quality-compute trade-off than training with previous datasets. Future work can use our EntityNet dataset to train CLIP models with all emergent properties much more efficiently, thus allowing for experiments, where training can be controlled. So far, this has been possible only with models of lacking quality.

\section*{Acknowledgements}
This work was funded by the Deutsche Forschungsgemeinschaft (DFG, German Research Foundation) – Project-ID 499552394 – SFB 1597 – Project-ID 417962828 – Project-ID 539134284. The authors acknowledge support from the state of Baden-Württemberg through bwHPC.

\setcounter{section}{0}
\renewcommand{\thesection}{\Alph{section}}

\section{Additional analyses}\label{sec:detailed-results}

\subsection{Verification of Bing search results}

In this work, we search for images that fit our queries using Bing image search. In contrast, CLIP~\cite{clip} searches for the query in a large pool of image-text pair candidates created from raw HTML via substring matching.
We manually evaluate the quality of the image search engine and of substring matching in \Cref{tab:verify-search}, using randomly select queries of our EntityNet dataset. In total, 87\% of queries were answered correctly.
Incorrect results of Bing search can be grouped into \emph{wrong similar terms}
(for the japanese ``dogi'' uniform, the search instead returns actual dogs),
and \emph{attribute ignored} (for ``B-25 Mitchell gray paint'' the search returns the correct B-25 plane, but differently painted).
Bing search is clearly superior to substring matching, with the latter algorithm only fulfilling 52\% of queries.

\subsection{Scaling behaviour of EntityNet}

In \Cref{fig:scaling}, We study the impact of removing entire entities or reducing the number of queries per entity, and compare this with randomly dropping images.
Removing entities and their associated images degrades performance more than randomly removing the same number of images, indicating that entities represent important semantic concepts for model training. In other words, our results suggest that diversity (more entities with fewer images per entity) is more important than depth (fewer entities with more images each).
In all cases, shrinking the dataset lowers performance, so a certain dataset size (here 33M) is indeed essential for a good generic model.

\section{Detailed experimental results}

We show extensive results on object classification in \Cref{tab:detailed-results-obj}.
For a more detailed analysis of model capabilities on ImageNet, we split the classes into \emph{living} (410 classes) and \emph{other} (590 classes) using WordNet:
Since ImageNet labels are built on WordNet nouns, we simply select all labels that are children of the \emph{living things} node for the \emph{living} set.
On iNaturalist, in addition to the 2021 version, we also evaluate on the 2019 version which contains 3,030 images in the validation set, each annotated with one of 1,010 species. We test with the same protocol as on iNaturalist 2021, testing on both english and latin class names and reporting the best accuracy.
We show additional results on retrieval in \Cref{tab:detailed-results-ret} and object classification under distribution shift in \Cref{tab:detailed-results-shift}.

\section{Experimental details}

\subsection{Hyperparameter settings} \label{sec:hparams}

We report hyperparameters for our experiments in \Cref{tab:hparams}.
Similar to Li~et~al.~\cite{li2023clipav2}, we reduce the context size of the text encoder down from 77 to 32 to reduce VRAM and training time requirements.
For a fair comparison with other CLIP models, we report all training cost and training duration as if the training was run at a context length of 77.

\subsection{Object classification evaluation}

To test the VLMs on object classification, we use the same procedure as CLIP~\cite{clip}:
Given an image $I$, class names $C_1, ..., C_{N}$, image encoder $f$ and text encoder $g$, we embed the image using the image encoder $\mathbf{v} = f(I)$. To acquire a text embedding for class $C_c$, the CLIP authors started by directly encoding the class names as $\mathbf{w}_c = g(C_c)$, e.g., \emph{dog}.
Alternatively, they created several prompts $P$ using templates, e.g., \emph{graffiti of a dog}, \emph{a photo of the cool dog}, etc., then encoded each prompt, and computed the average embedding:
$\mathbf{w}_c = \sum_{p \in P} g(p) / |P|$.
They referred to this approach as using ``context prompts''. Finally, given the image and text embeddings, the prediction $p$ is the class which has the highest cosine similarity to the image.

For a fair comparison between models that have been trained with different prompts,
we evaluate all models on just encoding the class name, and on using the average embedding of the 80 context prompts that the CLIP authors used for ImageNet, then report the higher top-1 accuracy.

\section{Qualitative example of text label sampling}\label{sec:dataset-example}

We show an example of our text label sampling strategy in \Cref{tab:text-label-sampling}.

\section{Image Search API details}\label{sec:search-apis}

\textbf{Google} The \href{https://developers.google.com/custom-search/v1/overview}{Google Image Search API} is available via the \href{https://cloud.google.com}{Google Cloud Platform}, and requires an existing \href{https://programmablesearchengine.google.com/}{programmable search engine} to function. It returns up to 10 images per request and page with a limit of 10 pages, i.e., 100 images per query.
It costs 5\$ per 1,000 API calls, resulting in costs of about 500\$ to download 1M images. We found the search results from the Google API to be quite different, and arguably worse, from the ones returned when using the regular \href{https://google.com/images}{Google image search}. For all our API requests we set the parameter \emph{imgColorType} to \emph{color}, \emph{imgType} to \emph{photo}, \emph{lr} to \emph{lang\_en}, and \emph{excludeTerms} to \emph{drawing clipart illustration cartoon vector painting}. This way we get mostly real-world images in our search results. We additionally add all aliases and the natural type of the sought entity to the \emph{orTerms} parameter for entity and entity-attribute queries. Because the Google API returns only up to 10 images per request and page, we search for the following number of pages: 2 pages each for entity queries, 4 pages each for entity-attribute queries, and 10 pages each for all natural-type-attribute queries.
We started our search with queries from the \organismsubset{} subset on both the Google and Bing APIs. We found the quality and value-for-money ratio of the Bing API to be better, and therefore switched to only using Bing for all other queries.

\textbf{Bing} The \href{https://www.microsoft.com/en-us/bing/apis/bing-image-search-api}{Bing Image Search API} is available via \href{https://azure.microsoft.com}{Microsoft Azure}. It returns up to 150 images per request and has no restrictions on the number of accessible pages. It costs 18\$ per 1,000 API calls, resulting in costs of about 120\$ to download 1M images. In our experience, the returned images closely match the ones from the regular \href{https://bing.com/images}{Bing image search}. For all our API requests we set the parameter \emph{imageType} to \emph{Photo} and \emph{color} to \emph{ColorOnly}. Unlike the Google API, Bing does not have a way to specify \emph{orTerms} via a separate request parameter, so we add the natural type of the sought entity to the search query directly, e.g., we search for \emph{jaguar animal}.
The Bing API returns 150 images per request and page, we request one page for each query. 

\section{Querying entities with SPARQL}\label{sec:sparql}

The SPARQL query over Wikidata used to harvest all entities under a super-entity is displayed in \Cref{fig:sparql-vehicle}. 
It returns a list of entities from a specified target domain as defined by one or more super-entities. The super-entities can be determined manually by searching for appropriate entities on the Wikidata website. For example, if we want to build a dataset about vehicles, we can set the super-entity to \href{http://www.wikidata.org/entity/Q1420}{vehicle (Q42889)}, as done in \Cref{tab:vehicle-entities}.
We list the super-entities we considered for our dataset and the relevant statistics in \Cref{tab:super-entities-used} for included and \Cref{tab:super-entities-skipped} for excluded super-entities.

\begin{table}[!htbp]
\centering
\caption{
We randomly select 100 search queries for each query set and manually check for the following errors: 
\emph{Wrong:} The majority of the results do not match the query.
\emph{Too few:} Only four or less images are found.
Over all 416k queries, Bing search finds $\ge 5$ images in 99.8\% of cases.
\postcaption{}
}
\label{tab:verify-search}
{
\tablefontsize
\setlength{\tabcolsep}{6pt}
\begin{tabular}{lrrrr}
\toprule
Query set & \multicolumn{2}{c}{Bing search} & \multicolumn{2}{c}{Substring matching} \\
& Wrong & Too few & Wrong & Too few \\
\midrule
World entity & 13\% & 0\% & 15\% & 26\% \\
World entity + attribute & 22\% & 0\% & 3\% & 66\% \\
Living entity & 0\% & 0\% & 7\% & 19\% \\
Living entity + attribute & 18\% & 0\% & 7\% & 51 \% \\
\midrule
Total & 13\% & 0\% & 8\% & 40 \% \\
\bottomrule
\end{tabular}
}
\end{table}

\begin{figure}[!htpb]
    \centering
    \includegraphics[width=0.69\linewidth]{fig-scaling.png}
    \caption{
Scaling entities and queries per entity.
    }
    \label{fig:scaling}
\end{figure}

\begin{table}[!htbp]
\caption{
Detailed object classification results.
\tblgroups{}
\tblexplainstar{}
\tblexplain{}
\tblexplainmacs{}
\tblexplainnotzs{}
\postcaption{}
}\label{tab:detailed-results-obj}
{
\fontsize{8pt}{8.648pt}\selectfont
\setlength{\tabcolsep}{2.4pt}
\begin{tabular}{lllrr|rrrrrrrrrrr}
\toprule
Arch.
& & Dataset
& MACs
& Images in
& \multicolumn{3}{c}{ImageNet}
& \multicolumn{2}{c}{iNat.}
& CUB & Rare
\\
& & & (1e18) & dataset (M) &
1k & living & other & 2019 & 2021 & & spcs \\
\multicolumn{5}{r}{\# Classes $\rightarrow{}$} & 1,000 & 410 & 590 & 1,010 & 10k & 200 & 400 \\
\midrule
B-32 & \ftno{} & \datasetcctwelvemshort{}  & 3.7 & 9.3 & 28.6 & 27.6 & 31.1 & 2.0 & 0.7 & 9.2 & \notzsm{} \\
B-32 & \ftno{} & CommonPool-M  & 2.9 & 128.0 & 27.2 & 20.6 & 33.5 & 2.6 & 0.8 & 10.1 & \notzsm{} \\
B-32 & \ftno{} & DataComp-M  & 2.9 & 14.0 & 29.7 & 25.5 & 34.5 & 3.0 & 1.0 & 16.8 & \notzsm{} \\
B-32 & \ftno{} & OpenAI  & 288.6 & 400.0 & \ranktwo{63.4} & 65.5 & \ranktwo{63.1} & 10.9 & 7.4 & 51.8 & \notzsm{} \\
B-32 & \ftno{} & DataComp-1B  & 295.4 & 1400.0 & \rankone{69.2} & \rankone{71.2} & \rankone{69.1} & \ranktwo{16.7} & \ranktwo{12.6} & \ranktwo{73.8} & \notzsm{} \\
\midrule
B-32 & \ftno{} & \entitynet{}  & 13.1 & 32.7 & 61.5 & \ranktwo{68.5} & 57.9 & \rankone{38.3} & \rankone{26.1} & \rankone{79.5} & \rankone{42.7} \\
\midrule
\midrule
B-16 & \ftyes{} & BioCLIP & 61.3 & 10.4 & 18.6 & 44.3 & 2.6 & \notzsm{} & \notzsm{} & 78.1 & \ranktwo{38.1} \\
B-16 & \ftno{} & CommonPool-L  & 78.2 & 1280.0 & 57.8 & 53.2 & 62.4 & 6.9 & 4.1 & 35.1 & \notzsm{} \\
B-16 & \ftno{} & DataComp-L  & 78.2 & 140.0 & 63.1 & 61.8 & 65.3 & 9.1 & 6.1 & 48.1 & \notzsm{} \\
B-16 & \ftno{} & DataComp-1B  & 791.4 & 1400.0 & \rankone{73.5} & \rankone{75.9} & \rankone{73.2} & \ranktwo{19.5} & \ranktwo{15.3} & \ranktwo{79.0} & \notzsm{} \\
B-16 & \ftno{} & OpenAI  & 784.6 & 400.0 & \ranktwo{68.3} & 71.5 & \ranktwo{67.4} & 12.5 & 9.2 & 56.1 & \notzsm{} \\
\midrule
B-16 & \ftno{} & \entitynet{}  & 36.0 & 32.7 & 66.2 & \ranktwo{73.9} & 62.0 & \rankone{42.2} & \rankone{32.0} & \rankone{85.3} & \rankone{47.1} \\
\midrule
\midrule
L-14 & \ftno{} & OpenAI  & 3328.4 & 400.0 & 75.5 & 78.9 & 74.5 & 15.2 & 12.0 & 62.9 & \notzsm{} \\
L-14 & \ftno{} & DataComp-1B  & 3338.6 & 1400.0 & 79.2 & 82.1 & 78.4 & 23.6 & 21.1 & 85.5 & \notzsm{} \\
L-14 & \ftno{} & DFN-2B  & 3338.6 & 2000.0 & \ranktwo{81.4} & \ranktwo{83.7} & \ranktwo{80.9} & 24.1 & 21.6 & \ranktwo{86.5} & \notzsm{} \\
H-14 & \ftno{} & DFN-5B  & 22164.0 & 5000.0 & \rankone{83.4} & \rankone{85.4} & \rankone{83.2} & \ranktwo{31.4} & \ranktwo{25.1} & \rankone{88.1} & \notzsm{} \\
\midrule
\midrule
B-32 & \ftno{} & DataComp-1B \ftbasemodel{} & 295.4 & 1400.0 & \ranktwo{69.2} & 71.2 & \rankone{69.1} & 16.7 & 12.6 & 73.8 & \notzsm{} \\
B-32 & \ftyes{} & \entitynet{}  & 13.3 & 32.7 & \rankone{69.5} & \ranktwo{73.6} & \ranktwo{67.8} & \ranktwo{41.5} & \ranktwo{29.5} & \ranktwo{83.3} & \notzsm{} \\
B-32 & \ftyes{} & \entityneta{}Only organisms  & 4.2 & 10.2 & 48.2 & \rankone{76.3} & 30.9 & \rankone{49.3} & \rankone{37.0} & \rankone{87.0} & \notzsm{} \\
\midrule
B-16 & \ftno{} & DataComp-1B \ftbasemodel{} & 791.4 & 1400.0 & \rankone{73.5} & 75.9 & \rankone{73.2} & 19.5 & 15.3 & 79.0 & \notzsm{} \\
B-16 & \ftyes{} & \entitynet{}  & 36.1 & 32.7 & \rankone{73.5} & \ranktwo{78.1} & \ranktwo{71.5} & \ranktwo{46.7} & \ranktwo{34.9} & \ranktwo{86.5} & \notzsm{} \\
B-16 & \ftyes{} & \entityneta{}Only organisms  & 11.3 & 10.2 & 51.4 & \rankone{80.2} & 33.8 & \rankone{54.3} & \rankone{42.7} & \rankone{90.3} & \notzsm{} \\
\midrule
\midrule
B-32 & \ftno{} & \entitynet{} & 13.1 & 32.7 & \rankone{61.5} & \rankone{68.5} & \rankone{57.9} & \ranktwo{38.3} & \ranktwo{26.1} & 79.5 & \rankone{42.7} \\
B-32 & \ftno{} & \entityneta{}No organisms  & 9.0 & 22.5 & 39.2 & 17.5 & \ranktwo{56.1} & 1.7 & 0.8 & 6.2 & 6.9 \\
B-32 & \ftno{} & \entityneta{}Only organisms  & 4.1 & 10.2 & 36.0 & \rankone{68.5} & 15.4 & \rankone{41.4} & \rankone{28.6} & \rankone{83.2} & \ranktwo{42.0} \\
B-32 & \ftno{} & \entityneta{}No attributes  & 8.7 & 21.8 & \ranktwo{54.8} & 63.0 & 50.4 & 36.4 & 25.6 & \ranktwo{79.7} & 39.2 \\
\midrule
B-32 & \ftno{} & \entityneta{}50\% alt text & 13.1 & 32.7 & \rankone{61.5} & \rankone{68.5} & \rankone{57.9} & \rankone{38.3} & \rankone{26.1} & \rankone{79.5} & \rankone{42.7} \\
B-32 & \ftno{} & \entityneta{}100\% alt text & 13.1 & 32.7 & \ranktwo{59.1} & \ranktwo{66.5} & \ranktwo{55.4} & \ranktwo{36.4} & 22.9 & \ranktwo{78.8} & \ranktwo{39.7} \\
B-32 & \ftno{} & \entityneta{}0\% alt text & 13.1 & 32.7 & 55.7 & 64.4 & 51.5 & 35.4 & \ranktwo{24.2} & 78.1 & 29.7 \\
\midrule
B-32 & \ftno{} & \entityneta{}Full size & 13.1 & 32.7 & \rankone{61.5} & \rankone{68.5} & \rankone{57.9} & \rankone{38.3} & \rankone{26.1} & \rankone{79.5} & \rankone{42.7} \\
B-32 & \ftno{} & \entityneta{}1/2 size  & 6.6 & 16.4 & \ranktwo{54.1} & \ranktwo{61.5} & \ranktwo{50.5} & \ranktwo{30.7} & \ranktwo{20.1} & \ranktwo{74.1} & \ranktwo{36.6} \\
B-32 & \ftno{} & \entityneta{}1/4 size  & 3.3 & 8.2 & 45.2 & 52.4 & 41.8 & 23.4 & 13.2 & 64.9 & 28.0 \\
B-32 & \ftno{} & \entityneta{}1/8 size  & 1.6 & 4.1 & 33.3 & 39.6 & 30.5 & 14.5 & 7.3 & 47.8 & 19.4 \\
B-32 & \ftno{} & \entityneta{}1/16 size  & 0.8 & 2.0 & 19.9 & 25.0 & 18.1 & 7.6 & 2.8 & 27.4 & 11.8 \\
\midrule
B-32 & \ftno{} & \entityneta{}Batch size 2k & 13.1 & 32.7 & 59.7 & 65.4 & 57.2 & 31.5 & 21.0 & 74.1 & 38.7 \\
B-32 & \ftno{} & \entityneta{}Batch size 4k & 13.1 & 32.7 & \ranktwo{60.9} & 67.0 & \rankone{58.0} & 36.2 & 24.1 & 77.6 & 41.0 \\
B-32 & \ftno{} & \entityneta{}Batch size 8k & 13.1 & 32.7 & \rankone{61.5} & \rankone{68.5} & \ranktwo{57.9} & 38.3 & 26.1 & 79.5 & \rankone{42.7} \\
B-32 & \ftno{} & \entityneta{}Batch size 16k & 13.2 & 32.7 & 60.5 & \ranktwo{68.2} & 56.6 & \ranktwo{38.8} & \rankone{26.9} & \ranktwo{81.2} & \ranktwo{42.1} \\
B-32 & \ftno{} & \entityneta{}Batch size 32k & 13.3 & 32.7 & 58.6 & 67.0 & 54.2 & \rankone{39.5} & \ranktwo{26.3} & \rankone{81.3} & 41.7 \\
\bottomrule
\end{tabular}
}
\end{table}

\begin{table}[!htbp]
\caption{
Detailed retrieval results.
\tblgroups{}
\tblexplainstar{}
\tblexplain{}
\tblexplainmacs{}
\postcaption{}
}\label{tab:detailed-results-ret}
{
\fontsize{8pt}{9.09pt}\selectfont
\setlength{\tabcolsep}{2.8pt}
\begin{tabular}{lllrr|r|rrrrrr}
\toprule
Arch.
& 
& Dataset
& MACs
& Images in
& Ave- & \multicolumn{2}{c}{COCO}
& \multicolumn{2}{c}{Flickr30K}
& \multicolumn{2}{c}{XM3600}
\\
& & & (1e18) & dataset (M) & rage & I2T & T2I & I2T & T2I & I2T & T2I
\\
\midrule
\midrule
B-32 & \ftno{} & \datasetcctwelvemshort{}  & 3.7 & 9.3 & 25.6 & 22.4 & 15.2 & 37.2 & 27.1 & 26.1 & 25.5 \\
B-32 & \ftno{} & CommonPool-M  & 2.9 & 128.0 & 20.2 & 18.3 & 11.2 & 29.9 & 18.9 & 23.6 & 19.6 \\
B-32 & \ftno{} & DataComp-M  & 2.9 & 14.0 & 19.5 & 17.1 & 11.0 & 26.0 & 18.0 & 23.6 & 21.5 \\
B-32 & \ftno{} & OpenAI  & 288.6 & 400.0 & \ranktwo{49.6} & \ranktwo{50.1} & \ranktwo{30.5} & \ranktwo{77.5} & \ranktwo{58.8} & \ranktwo{43.4} & 37.2 \\
B-32 & \ftno{} & DataComp-1B  & 295.4 & 1400.0 & \rankone{54.0} & \rankone{53.5} & \rankone{37.1} & \rankone{78.8} & \rankone{61.1} & \rankone{48.3} & \rankone{45.3} \\
\midrule
B-32 & \ftno{} & \entitynet{}  & 13.1 & 32.7 & 37.2 & 32.8 & 22.2 & 52.3 & 37.8 & 40.6 & \ranktwo{37.3} \\
\midrule
\midrule
B-16 & \ftyes{} & BioCLIP & 61.3 & 10.4 & 0.8 & 0.4 & 0.2 & 0.9 & 0.6 & 1.8 & 1.1 \\
B-16 & \ftno{} & CommonPool-L  & 78.2 & 1280.0 & 45.6 & 44.4 & 28.8 & 68.3 & 51.0 & 42.1 & 39.2 \\
B-16 & \ftno{} & DataComp-L  & 78.2 & 140.0 & 49.4 & 48.7 & 32.2 & 73.5 & 55.1 & \ranktwo{44.7} & \ranktwo{42.1} \\
B-16 & \ftno{} & DataComp-1B  & 791.4 & 1400.0 & \rankone{57.4} & \rankone{57.5} & \rankone{40.2} & \rankone{84.9} & \rankone{67.3} & \rankone{47.9} & \rankone{46.5} \\
B-16 & \ftno{} & OpenAI  & 784.6 & 400.0 & \ranktwo{52.1} & \ranktwo{52.5} & \ranktwo{33.1} & \ranktwo{81.9} & \ranktwo{62.0} & 43.8 & 39.3 \\
\midrule
B-16 & \ftno{} & \entitynet{}  & 36.0 & 32.7 & 39.8 & 36.0 & 25.2 & 57.1 & 43.3 & 39.9 & 37.1 \\
\midrule
\midrule
L-14 & \ftno{} & OpenAI  & 3328.4 & 400.0 & 54.3 & 56.3 & 36.5 & 85.1 & 65.2 & 44.5 & 38.4 \\
L-14 & \ftno{} & DataComp-1B  & 3338.6 & 1400.0 & 61.8 & 63.2 & 45.8 & 89.5 & 73.4 & 50.4 & 48.6 \\
L-14 & \ftno{} & DFN-2B  & 3338.6 & 2000.0 & \ranktwo{64.2} & \ranktwo{65.7} & \ranktwo{48.6} & \ranktwo{89.6} & \ranktwo{75.3} & \ranktwo{53.6} & \ranktwo{52.7} \\
H-14 & \ftno{} & DFN-5B  & 22164.0 & 5000.0 & \rankone{68.7} & \rankone{72.3} & \rankone{53.9} & \rankone{93.0} & \rankone{80.2} & \rankone{57.6} & \rankone{55.4} \\
\midrule
\midrule
B-32 & \ftno{} & DataComp-1B \ftbasemodel{} & 295.4 & 1400.0 & \rankone{54.0} & \rankone{53.5} & \rankone{37.1} & \rankone{78.8} & \rankone{61.1} & \rankone{48.3} & \ranktwo{45.3} \\
B-32 & \ftyes{} & \entitynet{}  & 13.3 & 32.7 & \ranktwo{50.8} & \ranktwo{48.1} & \ranktwo{34.4} & \ranktwo{72.1} & \ranktwo{57.1} & \ranktwo{47.8} & \rankone{45.6} \\
B-32 & \ftyes{} & \entityneta{}Only organisms & 4.2 & 10.2 & 31.2 & 28.0 & 19.7 & 44.8 & 33.7 & 30.9 & 29.9 \\
\midrule
B-16 & \ftno{} & DataComp-1B \ftbasemodel{} & 791.4 & 1400.0 & \rankone{57.4} & \rankone{57.5} & \rankone{40.2} & \rankone{84.9} & \rankone{67.3} & \rankone{47.9} & \rankone{46.5} \\
B-16 & \ftyes{} & \entitynet{}  & 36.1 & 32.7 & \ranktwo{52.2} & \ranktwo{50.4} & \ranktwo{36.4} & \ranktwo{75.6} & \ranktwo{59.4} & \ranktwo{47.3} & \ranktwo{44.3} \\
B-16 & \ftyes{} & \entityneta{}Only organisms  & 11.3 & 10.2 & 34.8 & 31.8 & 22.8 & 51.7 & 37.1 & 33.5 & 32.0 \\
\midrule
\midrule
B-32 & \ftno{} & \entitynet{} & 13.1 & 32.7 & \rankone{37.2} & \rankone{32.8} & \rankone{22.2} & \rankone{52.3} & \rankone{37.8} & \rankone{40.6} & \rankone{37.3} \\
B-32 & \ftno{} & \entityneta{}No organisms & 9.0 & 22.5 & \ranktwo{32.1} & \ranktwo{28.6} & \ranktwo{18.7} & \ranktwo{46.3} & \ranktwo{33.6} & \ranktwo{33.2} & \ranktwo{32.3} \\
B-32 & \ftno{} & \entityneta{}Only organisms & 4.1 & 10.2 & 16.5 & 12.8 & 9.8 & 23.4 & 15.3 & 19.0 & 18.6 \\
B-32 & \ftno{} & \entityneta{}No attributes & 8.7 & 21.8 & 28.6 & 23.6 & 16.1 & 41.3 & 27.3 & 32.3 & 30.8 \\
\midrule
B-32 & \ftno{} & \entityneta{}50\% alt text & 13.1 & 32.7 & \ranktwo{37.2} & \ranktwo{32.8} & \ranktwo{22.2} & \ranktwo{52.3} & \ranktwo{37.8} & \rankone{40.6} & \ranktwo{37.3} \\
B-32 & \ftno{} & \entityneta{}100\% alt text & 13.1 & 32.7 & \rankone{38.3} & \rankone{35.2} & \rankone{23.2} & \rankone{53.4} & \rankone{39.8} & \ranktwo{40.5} & \rankone{38.0} \\
B-32 & \ftno{} & \entityneta{}0\% alt text & 13.1 & 32.7 & 13.5 & 8.7 & 6.1 & 19.2 & 11.7 & 18.3 & 17.0 \\
\midrule
B-32 & \ftno{} & \entityneta{}Full size & 13.1 & 32.7 & \rankone{37.2} & \rankone{32.8} & \rankone{22.2} & \rankone{52.3} & \rankone{37.8} & \rankone{40.6} & \rankone{37.3} \\
B-32 & \ftno{} & \entityneta{}1/2 size  & 6.6 & 16.4 & \ranktwo{30.3} & \ranktwo{27.0} & \ranktwo{17.6} & \ranktwo{42.7} & \ranktwo{29.7} & \ranktwo{33.3} & \ranktwo{31.6} \\
B-32 & \ftno{} & \entityneta{}1/4 size  & 3.3 & 8.2 & 23.5 & 19.5 & 13.0 & 31.1 & 22.9 & 27.6 & 27.2 \\
B-32 & \ftno{} & \entityneta{}1/8 size  & 1.6 & 4.1 & 16.6 & 12.8 & 8.7 & 20.7 & 14.2 & 21.9 & 20.9 \\
B-32 & \ftno{} & \entityneta{}1/16 size  & 0.8 & 2.0 & 9.4 & 7.1 & 5.0 & 10.0 & 7.1 & 13.9 & 13.5 \\
\midrule
B-32 & \ftno{} & \entityneta{}Batch size 2k & 13.1 & 32.7 & 35.6 & 30.8 & 20.9 & 50.2 & \ranktwo{37.2} & 38.2 & 36.4 \\
B-32 & \ftno{} & \entityneta{}Batch size 4k & 13.1 & 32.7 & \ranktwo{36.4} & 31.6 & \ranktwo{22.1} & \ranktwo{51.7} & 37.0 & \ranktwo{39.1} & \ranktwo{37.2} \\
B-32 & \ftno{} & \entityneta{}Batch size 8k & 13.1 & 32.7 & \rankone{37.2} & \rankone{32.8} & \rankone{22.2} & \rankone{52.3} & \rankone{37.8} & \rankone{40.6} & \rankone{37.3} \\
B-32 & \ftno{} & \entityneta{}Batch size 16k & 13.2 & 32.7 & 35.8 & \ranktwo{32.2} & 21.8 & 50.5 & 36.9 & 37.2 & 36.1 \\
B-32 & \ftno{} & \entityneta{}Batch size 32k & 13.3 & 32.7 & 34.5 & 31.0 & 20.8 & 48.9 & 34.5 & 36.2 & 35.3 \\
\bottomrule
\end{tabular}
}
\end{table}

\begin{table}[!htbp]
\caption{
Detailed results on ImageNet distribution shifts.
\tblgroups{}
\tblexplainstar{}
\tblexplain{}
\tblexplainmacs{}
\emph{INet:} ImageNet.
\postcaption{}
}\label{tab:detailed-results-shift}
{
\fontsize{8pt}{8.93pt}\selectfont
\setlength{\tabcolsep}{2.35pt}
\begin{tabular}{lllrr|rrrrrrrrrr}
\toprule
Arch.
& 
& Dataset
& MACs
& Images in
& INet & Ave- & INet & INet
& Obj.-
& INet & INet
\\
& & & (1e18) & dataset (M) & 1K & rage & V2 & R & Net & Sketch & A \\
& & \multicolumn{2}{r}{\# Classes $\rightarrow{}$} & & 1,000 & --
& 1000 & 200 & 133 & 1000 & 200 \\  %
\midrule
B-32 & \ftno{} & \datasetcctwelvemshort{}  & 3.7 & 9.3 & 28.6 & 18.3 & 24.2 & 34.5 & 12.1 & 16.0 & 4.7 \\
B-32 & \ftno{} & CommonPool-M  & 2.9 & 128.0 & 27.2 & 19.8 & 22.5 & 33.0 & 20.9 & 18.4 & 4.3 \\
B-32 & \ftno{} & DataComp-M  & 2.9 & 14.0 & 29.7 & 20.5 & 24.4 & 34.0 & 19.7 & 19.3 & 4.9 \\
B-32 & \ftno{} & OpenAI  & 288.6 & 400.0 & \ranktwo{63.4} & \ranktwo{48.7} & \ranktwo{56.0} & \ranktwo{69.4} & \ranktwo{44.2} & 42.3 & \rankone{31.5} \\
B-32 & \ftno{} & DataComp-1B  & 295.4 & 1400.0 & \rankone{69.2} & \rankone{56.3} & \rankone{60.8} & \rankone{78.2} & \rankone{55.2} & \rankone{56.8} & \ranktwo{30.5} \\
\midrule
B-32 & \ftno{} & \entitynet{}  & 13.1 & 32.7 & 61.5 & 41.0 & 53.6 & 58.8 & 32.6 & \ranktwo{45.0} & 14.9 \\
\midrule
\midrule
B-16 & \ftyes{} & BioCLIP & 61.3 & 10.4 & 18.6 & 15.4 & 17.7 & 16.0 & 3.2 & 7.3 & 32.9 \\
B-16 & \ftno{} & CommonPool-L  & 78.2 & 1280.0 & 57.8 & 47.0 & 50.0 & 68.4 & 49.1 & 45.9 & 21.7 \\
B-16 & \ftno{} & DataComp-L  & 78.2 & 140.0 & 63.1 & 51.1 & 55.2 & 71.8 & 53.1 & \ranktwo{49.7} & 25.5 \\
B-16 & \ftno{} & DataComp-1B  & 791.4 & 1400.0 & \rankone{73.5} & \rankone{64.4} & \rankone{66.0} & \rankone{83.6} & \rankone{63.7} & \rankone{60.4} & \ranktwo{48.4} \\
B-16 & \ftno{} & OpenAI  & 784.6 & 400.0 & \ranktwo{68.3} & \ranktwo{58.6} & \ranktwo{61.9} & \ranktwo{77.7} & \ranktwo{55.3} & 48.2 & \rankone{49.9} \\
\midrule
B-16 & \ftno{} & \entitynet{}  & 36.0 & 32.7 & 66.2 & 47.4 & 59.2 & 64.1 & 40.9 & 48.9 & 23.9 \\
\midrule
\midrule
L-14 & \ftno{} & OpenAI  & 3328.4 & 400.0 & 75.5 & 71.4 & 69.9 & 87.9 & 69.0 & 59.6 & \rankone{70.7} \\
L-14 & \ftno{} & DataComp-1B  & 3338.6 & 1400.0 & 79.2 & \ranktwo{74.9} & 72.0 & \ranktwo{90.8} & \rankone{74.3} & 68.0 & 69.6 \\
L-14 & \ftno{} & DFN-2B  & 3338.6 & 2000.0 & \ranktwo{81.4} & 74.8 & \ranktwo{74.6} & 90.0 & \ranktwo{74.1} & \ranktwo{68.3} & 66.8 \\
H-14 & \ftno{} & DFN-5B  & 22164.0 & 5000.0 & \rankone{83.4} & \rankone{76.3} & \rankone{77.4} & \rankone{93.0} & 68.4 & \rankone{72.8} & \ranktwo{69.9} \\
\midrule
\midrule
B-32 & \ftno{} & DataComp-1B \ftbasemodel{} & 295.4 & 1400.0 & \ranktwo{69.2} & \rankone{56.3} & \ranktwo{60.8} & \rankone{78.2} & \rankone{55.2} & \rankone{56.8} & \rankone{30.5} \\
B-32 & \ftyes{} & \entitynet{}  & 13.3 & 32.7 & \rankone{69.5} & \ranktwo{53.3} & \rankone{61.9} & \ranktwo{74.2} & \ranktwo{47.9} & \rankone{56.8} & \ranktwo{25.6} \\
B-32 & \ftyes{} & \entityneta{}Only organisms & 4.2 & 10.2 & 48.2 & 33.3 & 43.1 & 56.7 & 19.1 & 32.1 & 15.8 \\
\midrule
B-16 & \ftno{} & DataComp-1B \ftbasemodel{} & 791.4 & 1400.0 & \rankone{73.5} & \rankone{64.4} & \ranktwo{66.0} & \rankone{83.6} & \rankone{63.7} & \rankone{60.4} & \rankone{48.4} \\
B-16 & \ftyes{} & \entitynet{}  & 36.1 & 32.7 & \rankone{73.5} & \ranktwo{61.0} & \rankone{66.5} & \ranktwo{79.0} & \ranktwo{56.6} & \ranktwo{59.8} & \ranktwo{42.9} \\
B-16 & \ftyes{} & \entityneta{}Only organisms & 11.3 & 10.2 & 51.4 & 39.2 & 46.1 & 62.1 & 25.9 & 35.1 & 26.8 \\
\midrule
\midrule
B-32 & \ftno{} & \entitynet{} & 13.1 & 32.7 & \rankone{61.5} & \rankone{41.0} & \rankone{53.6} & \rankone{58.8} & \rankone{32.6} & \rankone{45.0} & \rankone{14.9} \\
B-32 & \ftno{} & \entityneta{}No organism & 9.0 & 22.5 & 39.2 & 28.0 & 33.6 & 37.7 & \ranktwo{29.1} & 32.0 & 7.5 \\
B-32 & \ftno{} & \entityneta{}Only organisms & 4.1 & 10.2 & 36.0 & 21.0 & 31.5 & 39.6 & 8.1 & 17.6 & 8.1 \\
B-32 & \ftno{} & \entityneta{}No attributes & 8.7 & 21.8 & \ranktwo{54.8} & \ranktwo{33.8} & \ranktwo{47.9} & \ranktwo{49.2} & 24.4 & \ranktwo{36.9} & \ranktwo{10.5} \\
\midrule
B-32 & \ftno{} & \entityneta{}50\% alt text & 13.1 & 32.7 & \rankone{61.5} & \rankone{41.0} & \rankone{53.6} & \rankone{58.8} & \rankone{32.6} & \rankone{45.0} & \rankone{14.9} \\
B-32 & \ftno{} & \entityneta{}100\% alt text & 13.1 & 32.7 & \ranktwo{59.1} & \ranktwo{38.1} & \ranktwo{51.1} & \ranktwo{55.6} & \ranktwo{30.1} & \ranktwo{40.8} & \ranktwo{13.1} \\
B-32 & \ftno{} & \entityneta{}0\% alt text & 13.1 & 32.7 & 55.7 & 35.5 & 48.0 & 53.0 & 27.1 & 36.8 & 12.4 \\
\midrule
B-32 & \ftno{} & \entityneta{}Full size & 13.1 & 32.7 & \rankone{61.5} & \rankone{41.0} & \rankone{53.6} & \rankone{58.8} & \rankone{32.6} & \rankone{45.0} & \rankone{14.9} \\
B-32 & \ftno{} & \entityneta{}1/2 size  & 6.6 & 16.4 & \ranktwo{54.1} & \ranktwo{33.6} & \ranktwo{47.2} & \ranktwo{51.0} & \ranktwo{24.4} & \ranktwo{36.5} & \ranktwo{8.9} \\
B-32 & \ftno{} & \entityneta{}1/4 size  & 3.3 & 8.2 & 45.2 & 25.7 & 39.1 & 39.9 & 18.1 & 25.6 & 5.9 \\
B-32 & \ftno{} & \entityneta{}1/8 size  & 1.6 & 4.1 & 33.3 & 17.7 & 28.5 & 28.7 & 12.0 & 15.3 & 4.2 \\
B-32 & \ftno{} & \entityneta{}1/16 size  & 0.8 & 2.0 & 19.9 & 10.0 & 17.2 & 17.6 & 6.8 & 5.7 & 2.8 \\
\midrule
B-32 & \ftno{} & \entityneta{}Batch size 2k & 13.1 & 32.7 & 59.7 & 40.4 & 52.5 & \ranktwo{59.4} & 31.3 & 44.5 & 14.4 \\
B-32 & \ftno{} & \entityneta{}Batch size 4k & 13.1 & 32.7 & \ranktwo{60.9} & \rankone{41.0} & 53.0 & \rankone{60.0} & \ranktwo{32.1} & \ranktwo{44.9} & \rankone{14.9} \\
B-32 & \ftno{} & \entityneta{}Batch size 8k & 13.1 & 32.7 & \rankone{61.5} & \rankone{41.0} & \rankone{53.6} & 58.8 & \rankone{32.6} & \rankone{45.0} & \rankone{14.9} \\
B-32 & \ftno{} & \entityneta{}Batch size 16k & 13.2 & 32.7 & 60.5 & 39.7 & \ranktwo{53.2} & 58.0 & 30.6 & 42.9 & 13.8 \\
B-32 & \ftno{} & \entityneta{}Batch size 32k & 13.3 & 32.7 & 58.6 & 37.2 & 51.3 & 54.6 & 27.9 & 40.6 & 11.6 \\
\bottomrule
\end{tabular}
}
\end{table}

\begin{table}[!htbp]
\caption{Hyperparameters used for pretraining and finetuning, unless otherwise stated.
For all experiments we use the AdamW \cite{loshchilov2018adamw} optimizer with $\epsilon=1e-8$, $\beta_1=0.9$, $\beta2=0.98$.
\postcaption
}\label{tab:hparams}
{
\setlength{\tabcolsep}{4pt}
\begin{tabular}{llrrrrrrrr}
\toprule
Dataset & Model & Batch & Learning  & Weight & Epochs & Warmup \\
&& Size & Rate & Decay & & epochs  \\ \midrule
CC12M & ViT-B/32 & 8k & 5e-4 & 0.2 & 18 & 2 \\
CC12M & ViT-B/16 & 8k & 5e-4 & 0.2 & 18 & 2 \\
Ours, pretraining & ViT-B/32 & 8k & 5e-4 & 0.2 & 18 & 2 \\
Ours, pretraining & ViT-B/16 & 8k & 5e-4 & 0.2 & 18 & 2 \\
Ours, finetuning & ViT-B/32 & 32k & 5e-5 & 0.2 & 18 & 2 \\
Ours, finetuning & ViT-B/16 & 32k & 5e-5 & 0.2 & 18 & 2 \\ \bottomrule
\end{tabular}
}
\end{table}

\begin{table}[!htbp]
\caption{Example of our text label sampling strategy for an image returned from the entity query of \href{https://www.wikidata.org/wiki/Q101761}{zipper}. Probability mass is split 50/50 between image alt texts and texts from the knowledge graph. Between alt texts, we chose uniformly. Between knowledge graph texts we chose the search query 25\% of the time, a description 10\% of the time (uniformly between descriptions), and an alias otherwise (uniformly between all aliases).
\postcaption{}
}\label{tab:text-label-sampling}
\begin{minipage}[t]{0.68\textwidth}
\vspace{0pt}
{
\setlength{\tabcolsep}{4pt}
\renewcommand{\arraystretch}{1.3}
\begin{tabular}{
>{\raggedright}p{0.586\textwidth}|rr
}
\toprule
Text & Chance & Source \\
\midrule
Zipper PNG & 25\% & Alt text \\
yellow zipper PNG image & 25\% & Alt text \\
\midrule
zipper & 12.5\% & Search query \\
zip & 5.5\% & Alias \\
dingy & 5.5\% & Alias \\
clasp locker & 5.5\% & Alias \\
fly & 5.5\% & Alias \\
zip fastener & 5.5\% & Alias \\
device for fastening the edges of an opening of fabric or other flexible material & 2.5\% & Description \\
A device used for fastening, typically made of physical material. & 2.5\% & Description \\
\bottomrule
\end{tabular}
}
\end{minipage}
\hfill
\begin{minipage}[t]{0.3\textwidth}
\vspace{0pt}
\includegraphics[width=1\linewidth]{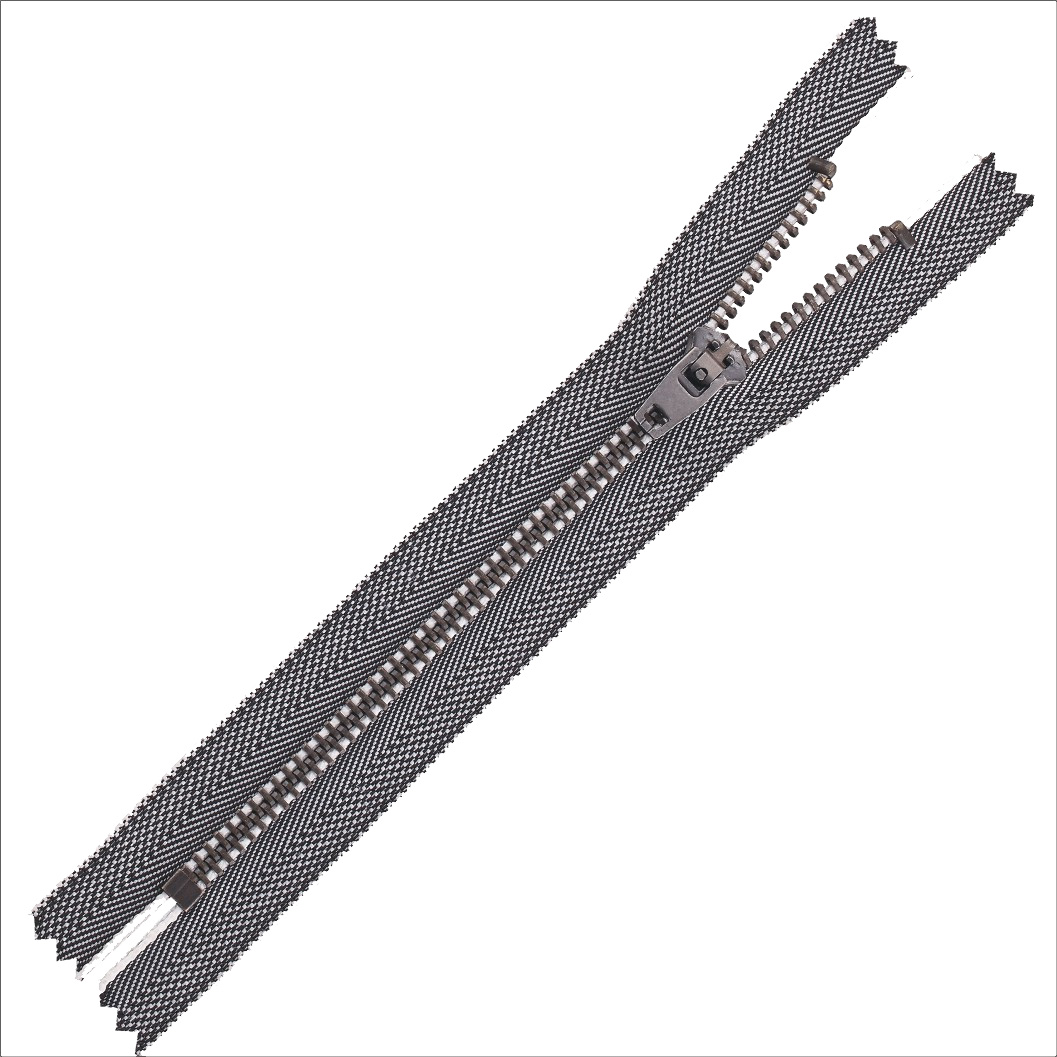}
\end{minipage}
\end{table}

\begin{figure}[!htbp]
\begin{myverbatim}
PREFIX wdt: <http://www.wikidata.org/prop/direct/>
PREFIX wd: <http://www.wikidata.org/entity/>
PREFIX rdfs: <http://www.w3.org/2000/01/rdf-schema#>
PREFIX schema: <http://schema.org/>
PREFIX wikibase: <http://wikiba.se/ontology#>
PREFIX skos: <http://www.w3.org/2004/02/skos/core#>
SELECT DISTINCT
  ?ent
  ?label
  ?desc
  ?links
  (GROUP_CONCAT(DISTINCT ?alias; SEPARATOR=";;;") AS ?aliases)
WHERE {
  VALUES ?typ { wd:Q42889 }
  ?ent wdt:P279* ?typ .
  ?ent rdfs:label ?label .
  FILTER(LANG(?label) = "en")
  ?ent ^schema:about/wikibase:sitelinks ?links .
  FILTER(?links >= 5)
  OPTIONAL {
    ?ent schema:description ?desc . 
    FILTER(LANG(?desc) = "en") 
  }
  OPTIONAL { 
    ?ent skos:altLabel ?alias . 
    FILTER(LANG(?alias) = "en") 
  }
}
GROUP BY ?ent ?label ?desc ?links
ORDER BY DESC(?links)
\end{myverbatim}
\caption{\
Generic SPARQL query for extracting entities from Wikidata that are related to a given set of super-entities. The super-entities are manually set within the \emph{VALUES ?typ \{ ... \}} clause. In this example it is the motor car entity wd:Q42889. A minimum number of sitelinks can also be specified to filter out unpopular entities, here it is set to 5.
}
\label{fig:sparql-vehicle}
\end{figure}

\begin{table}[!htbp]
\caption{
Vehicle entities and accompanying information as extracted from the Wikidata knowledge graph. Showing the first 5 and last 5 out of 17,015 entities. Note that we only collect entities with sitelinks $\ge 5$.
The corresponding SPARQL query is shown in \Cref{fig:sparql-vehicle}.
\postcaption{}
}\label{tab:vehicle-entities}
{
\setlength{\tabcolsep}{4pt}
\renewcommand{\arraystretch}{1.3}
\fontsize{9pt}{10pt}\selectfont
\begin{tabular}{
l
>{\raggedright}p{1.8cm}
>{\raggedright}p{2.8cm}l
>{\raggedright\arraybackslash}p{3.6cm}
}
\toprule
Identifier & Name & Description & Sitelinks & Aliases \\
\midrule
\href{http://www.wikidata.org/entity/Q1420}{Q1420} & motor car & motorized road vehicle designed to carry one to eight people rather than primarily goods & 237 & auto, motor vehicle, motor cars, motorcar, cars, car, automobiles, automobile, autocar \\
\href{http://www.wikidata.org/entity/Q11442}{Q11442} & bicycle & pedal-driven two-wheel vehicle & 203 & bike, Bicycles, cycle, pushbike, pedal cycle, pedal bike \\
\href{http://www.wikidata.org/entity/Q197}{Q197} & airplane & powered fixed-wing aircraft & 196 & airplane, aeroplane, plane, powered fixed-wing aircraft, planes, plane, aeroplane, fixed-wing powered aircraft, fixed-wing airplane, aeroplanes, fixed-wing aeroplane, airplanes \\
\href{http://www.wikidata.org/entity/Q870}{Q870} & train & form of rail transport consisting of a series of connected vehicles & 193 & rail-train, trains, railway train, railtrain, rail train, railroad train \\
\href{http://www.wikidata.org/entity/Q11446}{Q11446} & ship & large buoyant watercraft & 178 & marine vessel, vessel, water vessel, ships \\
\midrule
\href{http://www.wikidata.org/entity/Q813876}{Q813876} & Bedford JJL & motor vehicle & 5 &  \\
\href{http://www.wikidata.org/entity/Q7077241}{Q7077241} & Odakyu 20000 series RSE & Japanese electric multiple unit trainset & 5 & RSE, Romancecar RSE, Resort Super Express, Odakyu Romancecar RSE, 20000 series \\
\href{http://www.wikidata.org/entity/Q812263}{Q812263} & Bavarian Pt 2/3 & class of 97 German 2-4-0T locomotives & 5 & ÖBB 770, DR Class 70.0, DRG Class 70.0 \\
\href{http://www.wikidata.org/entity/Q9177196}{Q9177196} & Bombardier CRJ1000 & regional jet airliner & 5 & CRJ1000 \\
\href{http://www.wikidata.org/entity/Q812260}{Q812260} & Bavarian PtL 2/2 & class of 6+29+13 German 0-4-0T locomotives & 5 & DB Class 98.3, DRG Class 98.3, ÖBB 688 \\
\bottomrule
\end{tabular}
}
\end{table}

\begin{table}[!htbp]
\caption{
The super-entities for building our EntityNet dataset to describe the visual world.
The \emph{aliases} column refers to the set of all aliases collected from the entities. The numbers in this table are slightly higher than the ones we report in the main paper, because they refer to the raw counts of entities and aliases before profanity filtering and the removal of entities that return no results in the image search.
\postcaption{}
}\label{tab:super-entities-used}
\renewcommand{\arraystretch}{1.3}
\setlength{\tabcolsep}{4pt}
{\fontsize{7pt}{6.84pt}\selectfont
\begin{tabular}{p{1.5cm}p{4.8cm}p{2.6cm}rr}
\toprule
Super-entity & Description & Examples & Entities & Aliases \\
\midrule
product & Anything that can be offered to a market &
banh mi, navigation system%
& 63,676 & 144,715\\
substance & Any composed matter whose origin is either biological, chemical, or mineral &
solid lubricant, Chinese tea%
& 34,259 & 111,383\\
physical tool & Physical item that can be used to achieve a goal &
Patient lift, police transport%
& 32,727 & 71,227\\
animal & Kingdom of multicellular eukaryotic organisms &
saw-scaled viper, Sporathraupis cyanocephala%
& 28,000 & 76,408\\
plant & Living thing in the kingdom of photosynthetic eukaryotes &
Whitebark Pine, Eucalyptus coccifera%
& 28,000 & 55,925\\
material & Substance that can occur in different amounts, all with some similar [mixture of some] characteristics, and with which objects can be made &
dietary proteins, stone slab tomb%
& 18,021 & 40,822\\
vehicle & Mobile machine used for transport, whether it has an engine or not, including wheeled and tracked vehicles, air-, water-, and space-craft &
shipwrecks (objects), Evergreen A-class container ship%
& 17,015 & 37,849\\
geographical feature & Components of planets that can be geographically located &
hydrothermal Vents, grooves%
& 8,683 & 19,030\\
food & Any substance consumed to provide nutritional support for the body; form of energy stored in chemical 
&
coffee milk, tikka%
& 8,464 & 15,332\\
architectural structure & Human-designed and -made structure &
rock temples, summerhouse%
& 4,507 & 10,354\\
anatomical structure & Entity with a single connected inherent 3d shape that's created by coordinated expression of the organism's own dna &
bronchi, maxillary wisdom tooth%
& 4,394 & 9,999\\
facility & Place, equipment, or service to support a specific function &
public toilet, automobile servicing shop%
& 2,767 & 6,740\\
physical activity & Human physical activity consisting of voluntary bodily movement by skeletal muscles &
American rules football, archery%
& 2,228 & 4,422\\
clothing & Covering worn on the body &
blucher shoe, G-suit%
& 1,929 & 4,313\\
building & Structure, typically with a roof and walls, standing more or less permanently in one place &
shoestore, family restaurant%
& 1,655 & 3,964\\
musical instrument & Device created or adapted to make musical sounds &
electroencephalophone, Chinese flutes%
& 1,450 & 3,493\\
organ & Collection of tissues with similar functions &
nasal bone, cranial nerves%
& 1,155 & 2,450\\
furniture & Movable objects used to equip households, offices, or shops for purposes such as storage, seating, sleeping &
faldstool, airline seat%
& 388 & 933\\
body of water & Any significant accumulation of water, generally on a planet's surface &
dammed lake, deep-sea hydrothermal vent%
& 379 & 792\\
weather & State of the atmosphere &
cold snap, tropical cyclone%
& 151 & 304\\
precipitation & Liquid or solid water that falls to the ground &
hail, thunderstorm%
& 43 & 72\\
\midrule
Total & Before deduplication & & 259,891 & 620,527 \\
Total & After deduplication & & 146,985 & 368,062 \\
Total & \multicolumn{2}{l}{After deduplication, without animals and plants} & 90,985 & 235,795 \\
\bottomrule
\end{tabular}
}
\end{table}

\begin{table}[!htbp]
\caption{We consider these super-entities either non-visual, irrelevant, or too specific and do not select related entities when building our dataset.
\postcaption{}
}\label{tab:super-entities-skipped}
\renewcommand{\arraystretch}{1.3}
\setlength{\tabcolsep}{4pt}
{\fontsize{9pt}{10.00pt}\selectfont
\begin{tabular}{p{2.0cm}p{9.65cm}}
\toprule
Super-entity & Description\\
\midrule
abstract entity & entity that does not have a physical existence, including abstract objects and properties\\
astronomical object & physical body of astronomically-significant size, mass, or role, naturally occurring in a universe\\
city & large human settlement\\
concept & semantic unit understood in different ways, e.g. as mental representation, ability or abstract object (philosophy)\\
continent & large landmass identified by convention\\
country & distinct territorial body or political entity\\
historical event & particular incident in history that brings about a historical change\\
history & past events and their tracks or records\\
imaginary character & character known only from narrations (fictional or in a factual manner) without a proof of existence; includes fictional, mythical, legendary or religious characters and similar\\
language & particular system of communication, often named for the region or peoples that use it\\
language & structured system of communication\\
medical procedure & process of medicine done to heal; course of action intended to achieve a result in the delivery of healthcare\\
organization & social entity established to meet needs or pursue goals\\
planet & celestial body directly orbiting a star or stellar remnant\\
religion & social-cultural system\\
representation & entity or process that portrays something else, usually in a simplified or approximated manner\\
role & social role with a set of powers and responsibilities within an organization\\
science & systematic endeavor that builds and organizes knowledge, and the set of knowledge produced by this system\\
social system & patterned series of interrelationships existing between individuals, groups, and institutions\\
speciality & field limited to a specific area of knowledge; specialization in an occupation or branch of learning; a specific use\\
star & astronomical object consisting of a luminous spheroid of plasma held together by its own gravity\\
temporal entity & thing that can be contained within a period of time, or change in state (e.g. events, periods, acts)\\
work of art & aesthetic item or artistic creation; object whose value is its beauty only, not practical usefulness\\
written work & any work expressed in writing, such as inscriptions, manuscripts, documents or maps\\
\bottomrule
\end{tabular}
}
\end{table}

\clearpage
\bibliographystyle{splncs04_unsrt}%
\bibliography{083-main}

\end{document}